\documentclass[letterpaper, 10 pt, conference]{ieeeconf}  

\IEEEoverridecommandlockouts                              

\usepackage{amsmath} 
\usepackage{amssymb}  
\usepackage{xcolor}
\usepackage{hyperref}
\usepackage{booktabs}
\usepackage{tabularx}
\usepackage{xspace}
\usepackage{dblfloatfix} 
\usepackage{cuted}
\usepackage[font=footnotesize]{caption}

\newcommand{\ours}{\emph{DreamPlan}\xspace}

\newcommand{\eg}{\emph{e.g.}\xspace{}}

\newcommand{\figref}[1]{Fig.~\ref{#1}}
\newcommand{\tabref}[1]{Table~\ref{#1}}
\newcommand{\secref}[1]{Section~\ref{#1}}

\usepackage{tikzducks}

\title{\LARGE \bf
DreamPlan: Efficient Reinforcement Fine-Tuning of \\
Vision-Language Planners via Video World Models
}

\author{
   Emily Yue-Ting Jia$^{1,*}$\quad
   Weiduo Yuan$^{1,*}$\quad
   Tianheng Shi$^{1}$\quad
   Vitor Guizilini$^{2}$\quad
   Jiageng Mao$^{1,\dagger}$\quad
   Yue Wang$^{1,\dagger}$\\
   $^{1}$USC Physical Superintelligence Lab\quad $^{2}$Toyota Research Institute\\
   $^{*}$Equal contribution \quad
   $^{\dagger}$Equal advising
}

\begin{document}

\maketitle
\thispagestyle{empty}
\pagestyle{empty}

\begin{strip}
  \vspace{-12mm}
  \centering
  \includegraphics[width=\textwidth]{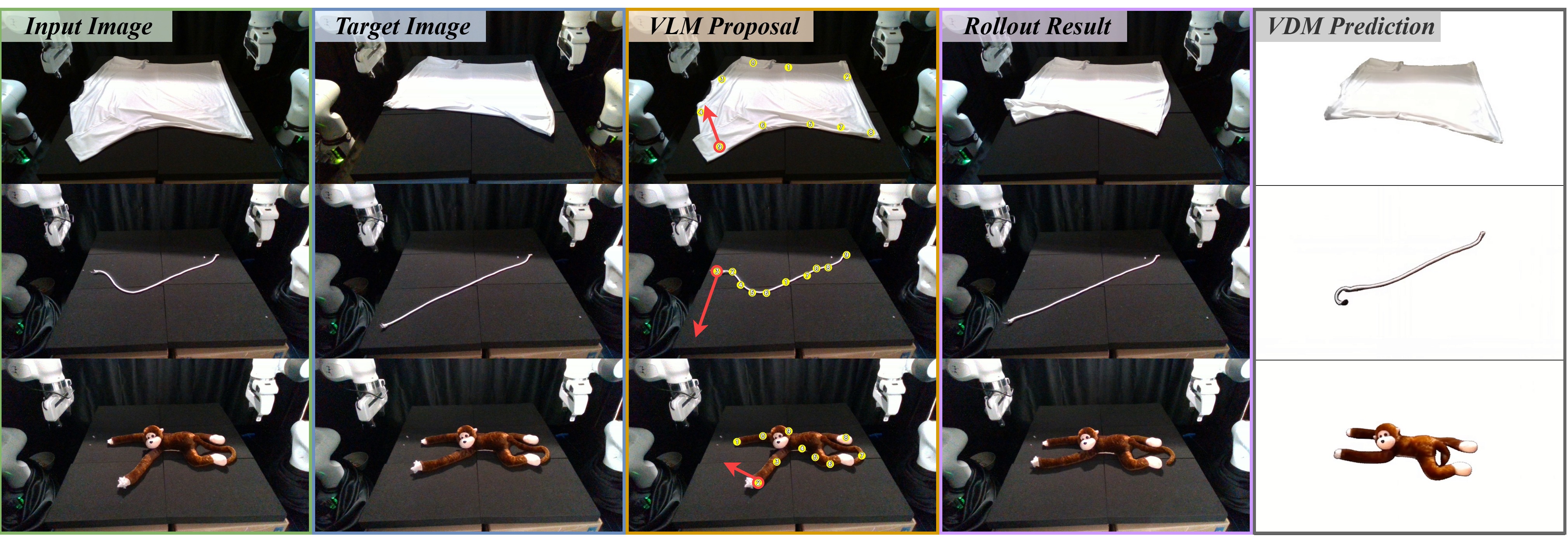}
  \captionof{figure}{\textbf{We propose \ours, a highly efficient framework that adapts vision-language (VLM) planners to real-world physics via virtual rollouts generated by world models.} \ours learns an action-conditioned video world model that captures task-specific deformable dynamics from exploratory interaction data collected by the zero-shot VLM, and leverages it to fine-tune the planner entirely offline. We demonstrate \ours's effectiveness on three challenging deformable manipulation tasks—cloth, rope, and soft toy manipulation—where DreamPlan significantly outperforms zero-shot baselines.}
  \label{fig:teaser}
  \vspace{-3mm}
\end{strip}

\begin{abstract}

Robotic manipulation requires sophisticated commonsense reasoning, a capability naturally possessed by large-scale Vision-Language Models (VLMs). While VLMs show promise as zero-shot planners, their lack of grounded physical understanding often leads to compounding errors and low success rates when deployed in complex real-world environments, particularly for challenging tasks like deformable object manipulation. Although Reinforcement Learning (RL) can adapt these planners to specific task dynamics, directly fine-tuning VLMs via real-world interaction is prohibitively expensive, unsafe, and sample-inefficient. To overcome this bottleneck, we introduce \ours, a novel framework for the reinforcement fine-tuning of VLM planners via video world models. Instead of relying on costly physical rollouts, \ours first leverages the zero-shot VLM to collect exploratory interaction data. We demonstrate that this sub-optimal data is sufficient to train an action-conditioned video generation model, which implicitly captures complex real-world physics. Subsequently, the VLM planner is fine-tuned entirely within the ``imagination" of this video world model using Odds Ratio Policy Optimization (ORPO). By utilizing these virtual rollouts, physical and task-specific knowledge is efficiently injected into the VLM. Our results indicate that \ours bridges the gap between semantic reasoning and physical grounding, significantly improving manipulation success rates without the need for large-scale real-world data collection. Our project page is \href{https://psi-lab.ai/DreamPlan/}{https://psi-lab.ai/DreamPlan/}.

\end{abstract}

\section{INTRODUCTION}

Robotic manipulation requires the ability to reason semantically and plan coherent multi-step behaviors. Large-scale Vision-Language Models (VLMs), trained on broad multimodal datasets, naturally possess strong semantic priors and reasoning ability, making them highly attractive as zero-shot planners. In this paradigm, a VLM can directly generate high-level plans, subgoals, or tool-use commands from visual observations and language instructions, which are subsequently executed by downstream low-level controllers. While this reasoning-centric approach demonstrates impressive generalization in simple, rigid-object scenarios with predictable dynamics, with contact-rich tasks that involve complex physical dynamics—most notably, the manipulation of deformable objects.

The primary bottleneck is that zero-shot VLM planners inherently lack task-specific physical knowledge. Consequently, the generated actions may be semantically plausible yet physically ineffective, as the model cannot reliably anticipate how objects will respond to contact, force, and gravity. This reality gap is severely magnified in deformable object manipulation, where object dynamics are complex, and minor deviations in action can lead to drastically different topological deformations. Although reinforcement learning (RL) offers a principled way to adapt VLM planners to specific task dynamics, directly fine-tuning these massive models via real-world robot interaction is prohibitively expensive, unsafe, and extremely sample-inefficient. 

To satisfy the massive interaction data requirements of RL, prior works mostly turn to simulation-based training~\cite{liu2023libero,li2024vla_rl}. However, constructing high-fidelity, computationally efficient simulators for deformable objects remains a tremendous challenge~\cite{lin2020softgym,muller2007position,yang2024maniskill3}, inevitably leading to a severe sim-to-real gap. More recently, video generation models have been explored as visual planners or implicit simulators~\cite{wu2024ctrlworld,wu2023daydreamer,xiao2025worldenv,zhu2025wmpo}. However, current action-conditioned visual models are largely limited to simple rigid-body interactions and struggle to realistically simulate complex deformable behaviors, limiting their usefulness for reliable policy fine-tuning.

To overcome these fundamental limitations, we propose \ours, a novel world-model-assisted RL framework designed to efficiently align VLM planners with real-world physics via model-generated virtual rollouts. \ours initiates the learning process by deploying the zero-shot VLM to collect exploratory interaction data in the target environment. Although the initial planner suffers from a low success rate—yielding a dataset dominated by failed or sub-optimal attempts—this data still contains dense, rich signals regarding action-outcome causality. We demonstrate that such sub-optimal exploratory data is surprisingly sufficient to train an action-conditioned video world model capable of predicting plausible object deformation. To effectively condition the video diffusion backbone on low-level robotic actions, we render the kinematic configuration of the robot arms corresponding to the commanded motions and apply a ControlNet-style architecture~\cite{Zhang_2023_controlnet} to inject these rendered visual action cues into the diffusion model. This allows the world model to accurately predict the physical evolution of the target objects under specific robot actions.

Empowered by the learned action-conditioned world model, \ours shifts policy adaptation entirely into the model’s imagination. Instead of relying on costly real-world interaction, we generate synthetic rollouts from the world model to provide virtual supervision for adapting the VLM planner. However, directly performing reinforcement learning with repeated video generation remains computationally expensive, as diffusion inference is inherently slow. To address this, we adopt an efficient Best-of-$K$ strategy: for each training input, we sample a batch of candidate actions from the planner and use the world model to predict their outcomes. The action that achieves the highest task-level objective under the predicted rollouts is selected as the positive sample, while the remaining actions are treated as negatives. These world-model-informed preferences are then used to fine-tune the planner via Odds Ratio Policy Optimization (ORPO), which increases the likelihood of preferred actions relative to disfavored ones without requiring repeated world-model queries during optimization,  enabling more efficient policy adaptation.

In summary, our main contributions are:
\begin{itemize}
    \item We introduce \ours, a highly efficient framework that adapts zero-shot VLM planners to complex physical tasks via offline reinforcement fine-tuning within video-generated virtual rollouts.
    \item We propose an efficient reinforcement adaptation strategy that integrates Best-of-$K$ sampling from the world model with Odds Ratio Policy Optimization (ORPO). This strategy decouples the computationally expensive video diffusion inference step from policy optimization, enabling highly efficient VLM adaptation with physics-informed supervision.
    \item We validate \ours on challenging deformable object manipulation tasks (\eg, cloth, ropes, and soft toys). Our results demonstrate that \ours significantly outperforms zero-shot baselines, efficiently injecting crucial physical grounding into VLMs without the heavy burden of large-scale real-world data collection.
\end{itemize}

\section{RELATED WORK}
\subsection{Vision-Language(-Action) Models for Robotics}
Recent works have extensively explored adapting pretrained Vision-Language Models (VLMs) and Vision-Language-Action (VLA) architectures for robotic manipulation. Works such as~\cite{ahn2022saycan,huang2022inner,driess2023palme} successfully align high-level semantic reasoning with robot execution via prompting, value grounding, or multimodal fine-tuning. However, these approaches typically rely on predefined skill primitives or require extensive expert demonstrations, limiting their ability to handle complex physical interactions.

To refine performance in specific tasks, recent frameworks leverage reinforcement learning (RL) for scalable policy post-training. Methods such as~\cite{li2024vlarl,intelligence2025pi06} use real interaction data or simulator-generated data to improve policy performance. Closely related to our approach, several emerging methods propose using learned world models as virtual environments to generate synthetic experience for policy fine-tuning. Works such as~\cite{guo2026vlaw,Liu2026worldvlaloop,xiao2025worldenv,zhu2025wmpo,mao2025robot,mao2025universal,zhao2025robot} integrate world models into the policy optimization loop, enabling imagined rollouts to reduce reliance on real-world interaction.

However, these methods primarily focus on rigid-body manipulation or relatively simple deformations, while we try to address the more challenging deformable manipulation task. Moreover, prior approaches typically perform heavy world model inference within the RL loop, whereas we introduce a more efficient Best-of-$K$ strategy that reduces video generation during policy updates while preserving physics-informed supervision.

\subsection{World Models for Robotics}
World models learn predictive environment dynamics and have been widely studied for improving data efficiency in reinforcement learning. Early works such as World Models~\cite{ha2018world} and Dreamer~\cite{hafner2019dream,hafner2021dreamerV2,hafner2023dreamerV3} showed that latent dynamics models enable policy optimization in imagination, substantially reducing real-world interaction, though they were primarily validated in simulation. Later extensions, such as DayDreamer~\cite{wu2023daydreamer}, applied this paradigm to real-world robotics by learning latent world models from physical interaction data. More recently, generative video models have been explored as world modeling tools due to their strong visual priors and temporal consistency. Methods such as Genie~\cite{bruce2024genie} and Ctrl-World~\cite{wu2024ctrlworld} build controllable video-based world models, while World-Env~\cite{xiao2025worldenv} and WMPO~\cite{zhu2025wmpo} leverage learned world models for VLA post-training. 

However, most existing approaches focus on rigid-body interactions or are trained mostly on demonstration datasets~\cite{ai2025review_dynamics,arriola2020modeling_deformable,gu2023survey_dom,zhao2025humanoid,jia2025learning,gao2025seeing,ye2025anchordreamrepurposingvideodiffusion}. Modeling more complex deformable dynamics remains underexplored. To this end, we train a task-specific action-conditioned video world model using interaction data collected on deformable objects and integrate it with reinforcement fine-tuning of a VLM planner.

\section{METHOD}
\begin{figure*}[t!]
    \centering
    \includegraphics[width=0.95\linewidth]{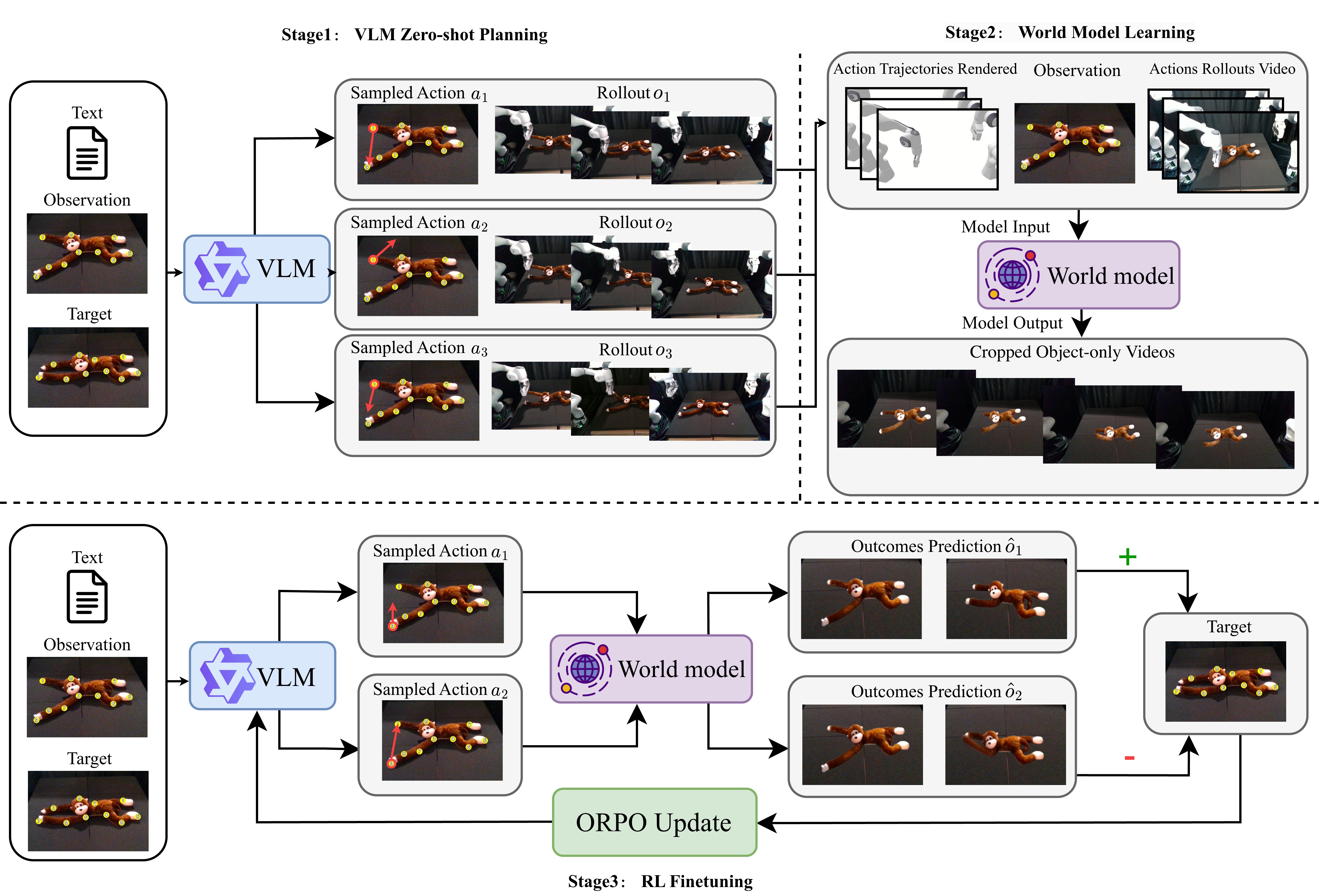}
    \caption{
    \textbf{Overview of \ours.}
    Our framework consists of three stages.
    \textbf{(1) Zero-shot proposal:} given the current observation and goal image, a pretrained VLM planner generates multiple candidate keypoint-based manipulation actions.
    \textbf{(2) World model learning:} these zero-shot actions are executed to collect diverse action–observation trajectories, which are used to fine-tune an action-conditioned diffusion world model that predicts object deformation outcomes from rendered robot-motion videos.
    \textbf{(3) World-model-guided alignment:} the trained world model acts as a verifier to evaluate sampled VLM actions by predicting their future outcomes; comparing predicted outcomes yields pairwise preferences (more vs. less goal-consistent actions), which are used to fine-tune the VLM planner via Odds Ratio Policy Optimization (ORPO), aligning it toward physics-consistent behaviors without additional real-world interaction.
    }
    \label{fig:method_overview}
    \vspace{-6mm}
\end{figure*}
We aim to tackle goal-conditioned deformable object manipulation, where a robot must transform an observed initial state into a desired target state specified by a goal image. At time step $ t $, the robot observes a visual state $o_t \in \mathcal{O}$ and is provided with a goal image $g \in \mathcal{O}$. The objective is to generate a sequence of actions $a_{0:T-1} \in \mathcal{A}$ that drives the environment toward the goal states. Solving such tasks requires not only high-level reasoning about spatial relationships, but also an implicit understanding of the underlying, highly nonlinear physical dynamics of deformable objects, governed by the transition distribution $o_{t+1} \sim \mathcal{P}(o_{t+1} \mid o_t, a_t)$.

Pretrained Vision-Language Models (VLMs) possess strong semantic priors and broad commonsense reasoning capabilities, making them attractive as zero-shot planners.  We therefore employ a VLM planner $\pi_\theta(a_t \mid o_t, g)$, parameterized by $\theta$, to generate actions conditioned on the current observation and goal. However, these foundational VLMs inherently lack task-specific physical grounding. To bridge this reality gap, we employ reinforcement learning (RL) to adapt the VLM to the target physical dynamics.
Directly collecting the massive scale of real-world interaction data typically required for RL is prohibitively expensive. To enable scalable adaptation, we instead learn an action-conditioned video world model $ W_\phi $, parameterized by $ \phi $, that approximates the environment dynamics: $ \hat{o}_{t+1:t+H} = W_\phi(o_t, g, a_{t:t+H-1})$, where $ H $ denotes the rollout horizon. Built upon a pretrained video diffusion backbone and fine-tuned on data $D = \left\{ \left(o_0^i, g^i, a_{0:H-1}^i, o_{1:H}^i \right) \right\}_{i=1}^{N_D}$ collected from zero-shot executions,  $ W_\phi $ captures task-specific action–outcome relationships for deformable manipulation. By substituting the true dynamics $ \mathcal{P} $ with our learned world model $ W_\phi $, we generate synthetic rollouts that serve as a scalable, offline training environment for reinforcement fine-tuning of the VLM planner $\pi_\theta$. The overall pipeline of our framework, \ours, is illustrated in ~\figref{fig:method_overview}.

\subsection{VLM Planner for Deformable Manipulation}
\label{subsection:vlm_planner}
Pretrained Vision-Language Models (VLMs) possess strong commonsense reasoning and spatial understanding, making them highly attractive as zero-shot planners for goal-conditioned manipulation. To effectively leverage these capabilities for complex deformable objects, we design a structured visual prompting strategy that constrains the action space while preserving the model's reasoning flexibility.

Given the current observation $o_t \in \mathcal{O}$ and a goal image $g \in \mathcal{O}$, we extract two sets of visual keypoints: $\mathcal{K}_t = \{k_{t}^i\}_{i=1}^{N_t}$ from $o_t$, and $\mathcal{K}_g = \{k_{g}^j\}_{j=1}^{N_g}$ from $g$. The keypoints are detected independently in the two images and are not explicitly matched across views. We plot these detected keypoints onto both images and provide them as visual context to the VLM planner $\pi_\theta(a_t \mid o_t, g)$.

Rather than regressing continuous pixel coordinates, the planner outputs a discrete manipulation command that selects a source keypoint $k_{s}^i \in \mathcal{K}_s$ to define the grasp location, and a target keypoint $k_{g}^j \in \mathcal{K}_g$ to specify the intended placement location. Accordingly, each action is defined as:
\begin{equation}
    a_t = (k_s^{i_t}, k_g^{j_t}),
\end{equation}
This discrete, keypoint-based action representation forces the planner to ground its reasoning in meaningful regions, drastically reducing spatial ambiguity and simplifying the optimization for subsequent reinforcement fine-tuning.

\subsection{Learning World Model for VLM Fine-Tuning}
\label{subsection:vdm_verifier}
The pretrained VLM planner lacks task-specific physical grounding, making reinforcement learning (RL) necessary to adapt it to deformable dynamics. However, large-scale real-world RL is prohibitively time-consuming and costly. To avoid costly real-world interaction, we construct an action-conditioned video world model $W_\phi$ to approximate the deformable dynamics $\mathcal{P}$ and provide a virtual training environment for policy refinement.

To support reinforcement fine-tuning in imagination, we need an action-conditioned video world model $W_\phi$ that captures the complex dynamics $\mathcal{P}$ of deformable objects. While prior works~\cite{wu2024ctrlworld,wu2023daydreamer,xiao2025worldenv,zhu2025wmpo} have explored action-conditioned world models for robotics, they are predominantly trained on rigid-body interactions—which exhibit significantly simpler dynamics—or on heavily curated successful demonstrations that inherently limit action diversity. Consequently, these existing models are inadequate for our deformable manipulation tasks, where the physical dynamics are highly nonlinear and exquisitely sensitive to subtle variations in action trajectories. To overcome this, we train a task-specific world model using exploratory data collected from the zero-shot VLM. While this dataset is dominated by sub-optimal executions, it provides dense, rich supervision over action-outcome causality, capturing the diverse distributions of physical interactions necessary for robust dynamic modeling.

A central challenge in training the world model lies in effectively conditioning the video diffusion model (VDM) on low-level robotic actions while simultaneously forcing the network to focus on object deformation rather than irrelevant visual details. In deformable manipulation, even minor variations in robot motion can significantly alter the pixel-level appearance of the object. To address this, we render the robot arm configurations corresponding to each action sequence, producing synthesized videos of the intended motions. We then inject these rendered action videos into the VDM using a ControlNet architecture, whose structural conditioning enforces strong pixel-level alignment between the action signals and the generated videos. Specifically, given the data $D = \left\{ \left(o_0^i, g^i, a_{0:H-1}^i, o_{1:H}^i \right) \right\}_{i=1}^{N_D}$ collected from zero-shot executions, the model predicts noise for the $i$-th data as:


\begin{equation}
    \hat{\epsilon} = \epsilon_\theta(\mathbf{x}^i_t, t, c^i) 
    + \Delta_\phi(\mathbf{x}^i_t, t, \mathbf{r^i}),
\end{equation}
where $c^i$ is the text and input image conditions, $\epsilon_\theta$ denotes the frozen pretrained backbone and $\Delta_\phi$ is the residual control branch that injects structured action features into corresponding layers of the diffusion network. $\mathbf{r^i}$ is the rendered robot arm trajectory corresponding to the action sequence:
\begin{equation}
    \textbf{r}^i = \text{render}(a_{0:H-1}^i),
\end{equation}
The predicted noise is optimized with: 
\begin{equation}
    \mathcal{L}_{\text{diff}} = 
    \mathbb{E}_{\mathbf{x}^i_0, t, \epsilon}
    \left[
    \|\epsilon - \hat{\epsilon}(\mathbf{x}^i_t, t, c, \mathbf{r}^i)\|_2^2
    \right],
\end{equation}

Additionally, to reduce the burden of reconstructing irrelevant visual details such as lighting, robot arms and background clutter, we train the model to predict cropped object-only videos with a white background, allowing it to concentrate on deformation dynamics.

In practice, we adopt CogVideoX-5B (image-to-video)~\cite{yang2024cogvideox} as the diffusion backbone and fine-tune it on trajectories collected from zero-shot executions. The dataset consists of several hundred action sequences, collected in approximately four hours of robot interaction. Despite the relatively modest data scale, this exploratory data proves sufficient to train a stable action-conditioned world model capable of generating plausible deformation rollouts for downstream reinforcement learning.

\begin{figure*}[t!]
    \centering
    \includegraphics[width=\linewidth]{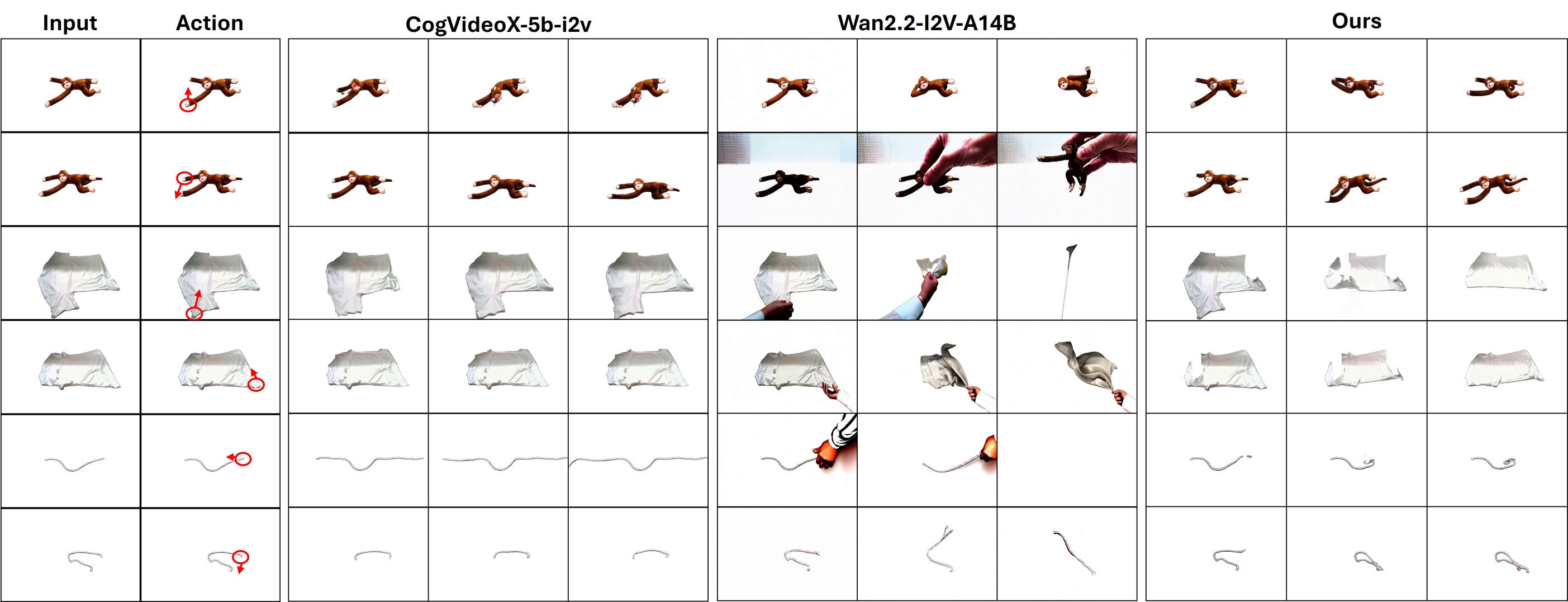}
    \caption{\textbf{Qualitative comparison of video generation results.} Baselines fail to follow the specified actions or produce unrealistic deformations, while our method generates deformations that is both action-consistent and physically plausible, demonstrating its reliability as a verifier for VLM fine-tuning.}
    \label{fig:vdm_qualitative}
    \vspace{-6mm}
\end{figure*}

\subsection{Reinforcement Fine-Tuning of VLM Planner with World Model}
With the learned action-conditioned world model $W_\phi$ successfully capturing the complex environment dynamics $\mathcal{P}$, we can shift policy adaptation entirely into the model's imagination, enabling reinforcement learning to specialize the VLM planner to task-specific physical dynamics without time-consuming real-world rollouts. Specifically, for a given observation-goal pair $(o_t, g)$, we aim to generate synthetic rollouts using $W_\phi$ to serve as virtual experience for fine-tuning the VLM planner $\pi_\theta(a_t \mid o_t, g)$ via reinforcement learning. However, direct online RL within this generative framework remains computationally expensive. In practice, a single video diffusion inference for one action sequence can take approximately one minute, making repeated world-model queries inside the optimization loop time-consuming.

To bypass this computational barrier, we adopt an efficient two-stage reward construction scheme based on Best-of-$K$ sampling. For each training pair in the collected data $(o^i_0, g^i) \in D$, we first sample a batch of candidate action sequences $\{a_i\}_{i=1}^K$ from the VLM planner. We then query the world model to predict their corresponding physical outcomes $\{\hat{o}_i\}_{i=1}^K$. Based on a task-level objective computed on the predicted rollouts, we select the optimal action $a^*$ as the positive sample, while the remaining actions in the pool are treated as negative samples. This transforms the physics-informed evaluation from the world model into structured preference signals without requiring further video generation during policy updates.

We optimize the planner using Odds Ratio Policy Optimization (ORPO), a preference-based fine-tuning method that directly contrasts positive and negative responses without requiring a separate value function. Given an input $s^i = (o_0, g^i)$, the planner produces a positive action $a^*$ and a set of negative actions ${a_j^{-}}$. ORPO encourages the policy to increase the likelihood of the positive action relative to the negatives by optimizing a contrastive objective based on the log-likelihood ratio:
\begin{equation}
    \mathcal{L}_{\text{ORPO}} =
\log \sigma \left(\log \pi_\theta(a^*|s^i) - \log \pi_\theta(a^{-}|s^i)\right),
\end{equation}
where $\sigma(\cdot)$ denotes the sigmoid function and $a^{-}$ represents sampled negative actions. This objective directly maximizes the odds of preferred actions over disfavored ones, providing a stable and memory-efficient alternative to traditional policy-gradient RL.

In our implementation, we use Qwen3-VL-8B~\cite{bai2025qwen3vl} as the backbone for the planner and GPT-4O~\cite{openai2024gpt4ocard} to select the best action by comparing predicted rollout images with the goal image $g$. This optimization strategy substantially reduces world-model inference overhead while preserving physics-informed supervision for efficient VLM planner fine-tuning.

\begin{figure}[t!]
    \centering
    \includegraphics[width=\linewidth]{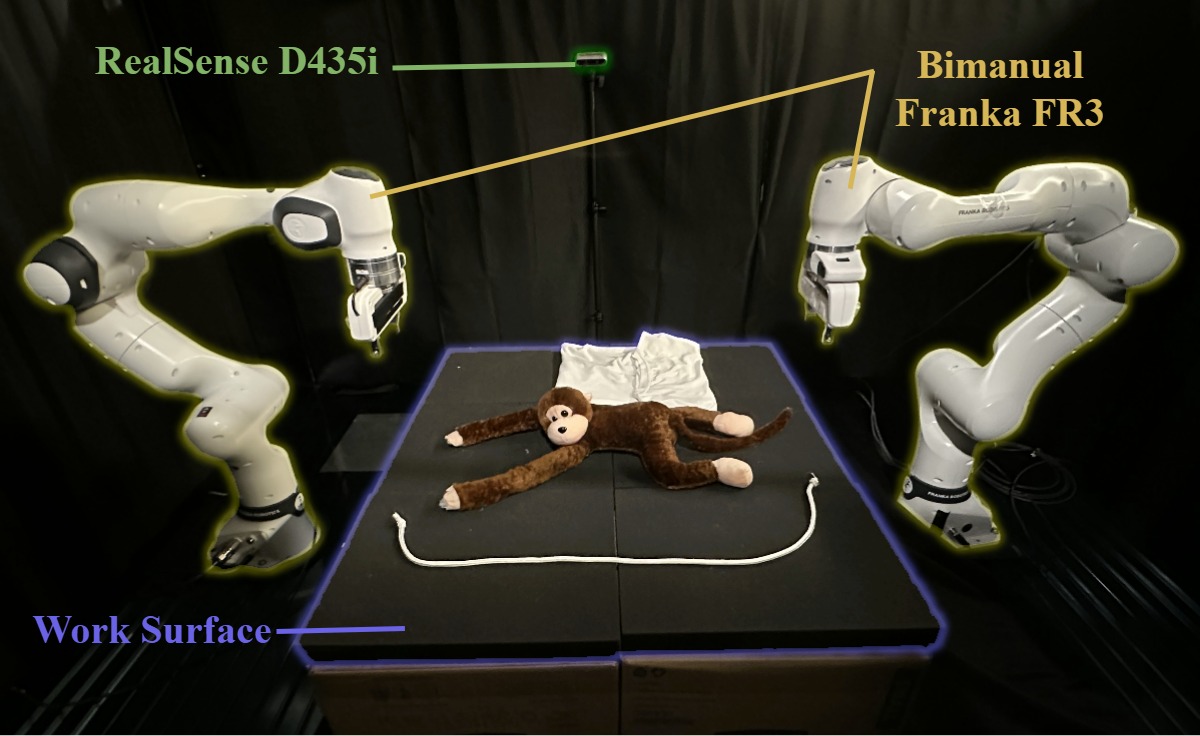}
    \caption{\textbf{Hardware setup.} Our real-world platform consists of two Franka FR3 arms positioned opposite each other to enable bimanual manipulation within a shared workspace. A top-mounted RealSense D435i camera captures the interaction area, where deformable objects are placed on the work surface for automated data collection and further manipulation.}
    \label{fig:hardware_setup}
    \vspace{-15pt}
\end{figure}

\section{EXPERIMENTS}

In this section, we investigate the following two questions through extensive experiments:

\textbf{(Q1)} Can video world models effectively simulate physical dynamics and serve as reliable verifiers for planning? 
(Section \textcolor{red}{\ref{subsec:vdm_as_verifier}})

\textbf{(Q2)} Can our reinforcement learning framework enable the VLM to internalize the world model's predictive understanding of physical dynamics? 
(Section \textcolor{red}{\ref{subsec:wm_rl}})

\begin{table*}[b!]
    \centering
    \newcolumntype{C}{>{\centering\arraybackslash}X}
    \caption{\textbf{Quantitative results on real-world tasks.} 
    Through \ours RL fine-tuning, Qwen3-VL-8B achieves state-of-the-art (SOTA) performance in complex object manipulation, consistently outperforming leading pure zero-shot baselines. 
    See~\secref{subsubsec:evaluation_protocol} for our evaluation protocol.}
    \label{tab:rl_vs_zeroshot}
    \begin{tabularx}{\linewidth}{lCCCc}
        \toprule
        \textbf{Method} & \textbf{Rope Straightening $\uparrow$} & \textbf{Cloth Folding $\uparrow$} & \textbf{Toy Arm Repositioning $\uparrow$} & \textbf{Avg score $\uparrow$}\\
        \midrule
        Qwen3-VL-4B           & 0.10 & 0.15 & 0.40 & 0.22  \\
        Qwen3-VL-8B           & 0.20 & 0.10 & 0.70 & 0.33  \\
        Qwen3-VL-32B          & 0.30 & 0.05 & 0.70 & 0.35  \\
        GPT-4o                & 0.30 & 0.00 & 0.55 & 0.28  \\
        \midrule
        Qwen3-VL-8B + \ours RL finetune       & \textbf{0.60} & \textbf{0.35} & \textbf{0.85} & \textbf{0.60} \\
        \bottomrule
    \end{tabularx}
\end{table*}

\subsection{Real-world experiments}
\subsubsection{Hardware Setup}
Our real-world experiments are conducted on a bimanual manipulation system, as illustrated in ~\figref{fig:hardware_setup}, consisting of two Franka FR3 arms~\cite{fr3}, each equipped with a standard Franka hand. The robots are mounted opposite to each other, positioned 140 cm. For visual observations, we use a mounted third-view RealSense D435i camera to record 960$\times$540 RGB images at 30 Hz. Both robotic arms are operated under Cartesian end-effector pose control at a frequency of 30 Hz. The entire framework is deployed locally on a workstation equipped with a 32GB NVIDIA RTX 5090 GPU.

\begin{figure}[t!]
    \centering
    \includegraphics[width=\linewidth]{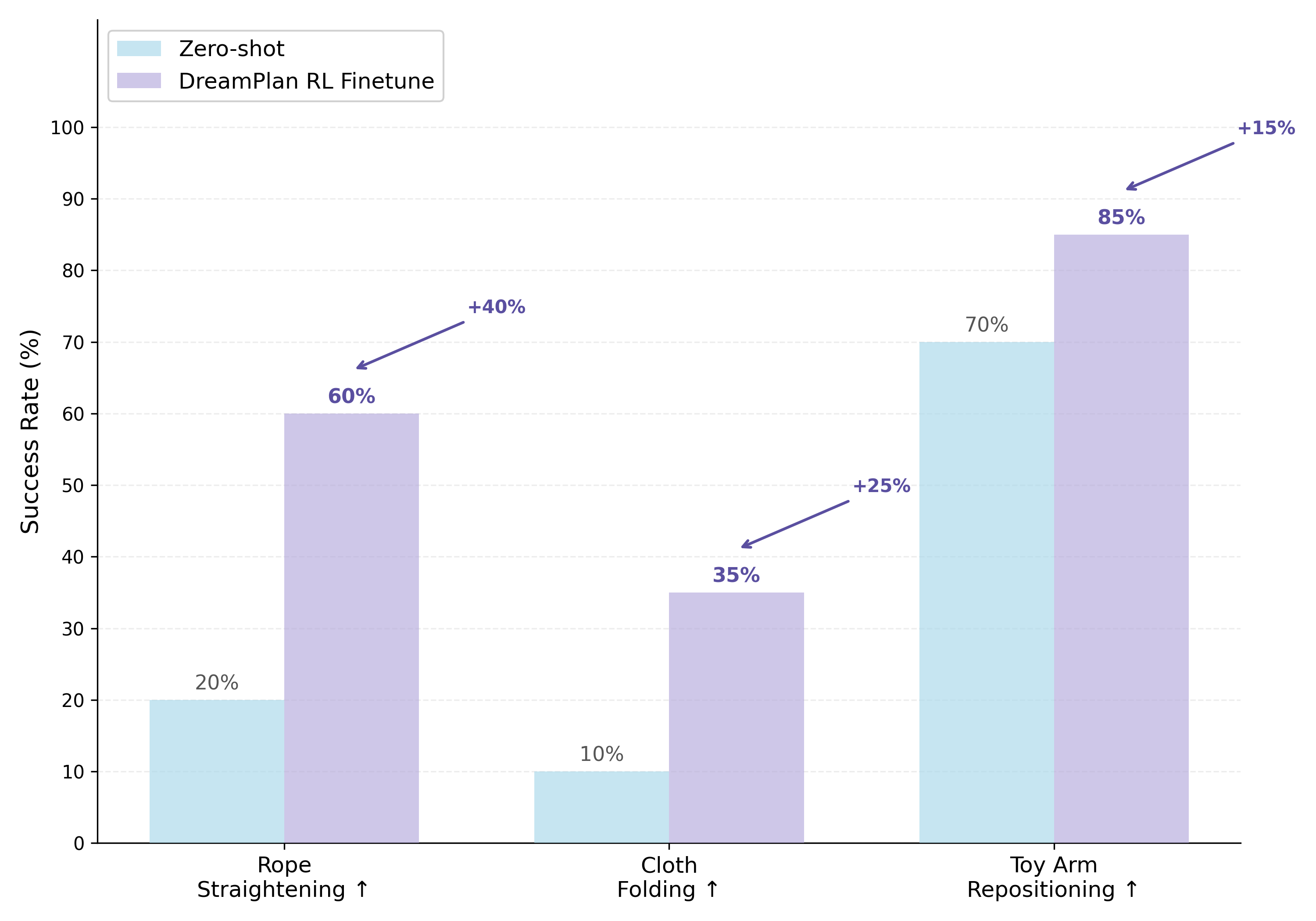}
    \caption{\textbf{Success rate improvements via RL fine-tuning.} Our RL pipeline improves success rates by 15\%--40\% over zero-shot baselines across all evaluated real-world tasks.}
    \label{fig:zeroshot_vs_rl}
    \vspace{-15pt}
\end{figure}
\begin{figure*}[t!]
    \vspace{1mm}
    \centering
    \includegraphics[width=0.76\linewidth]{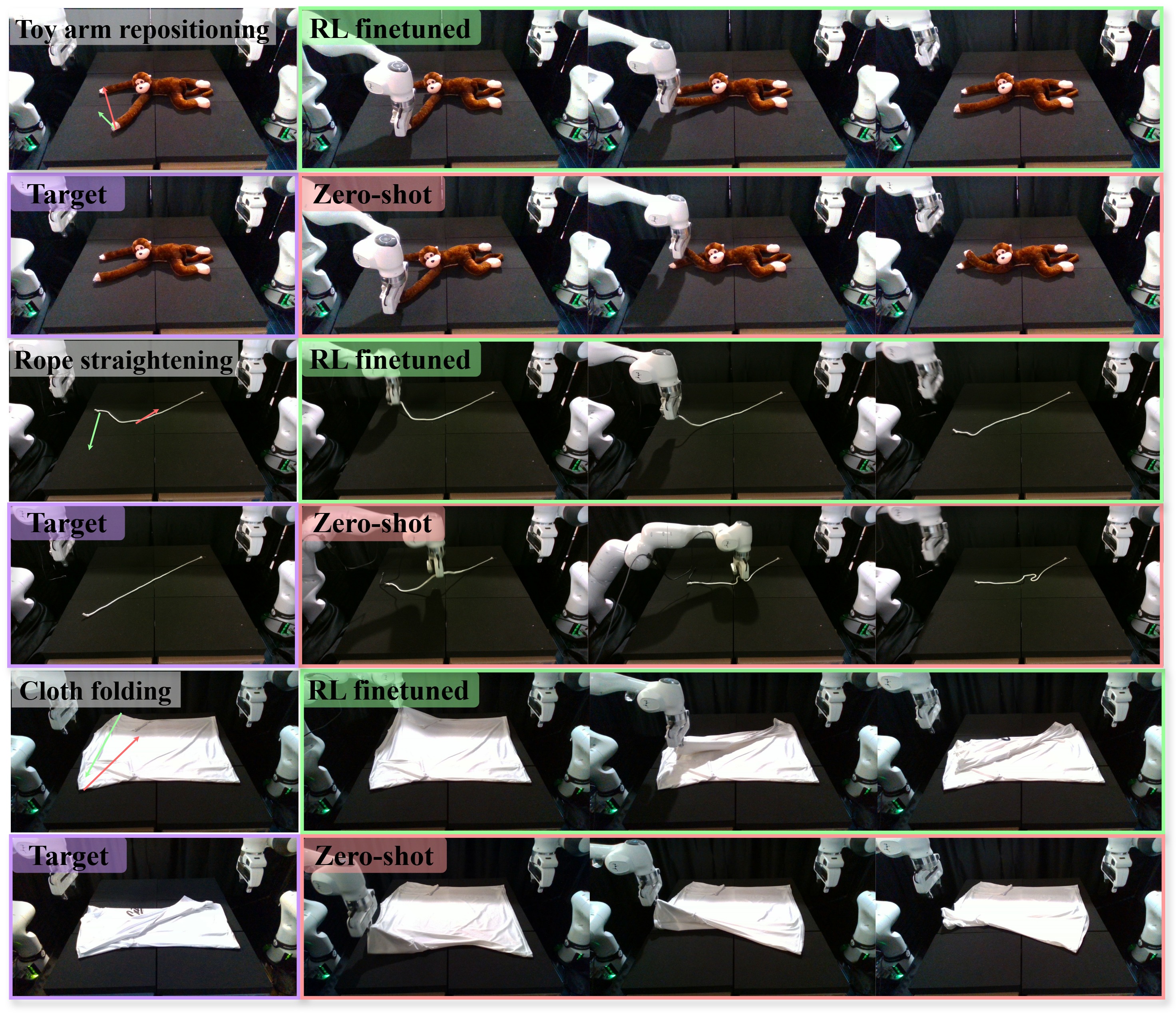}
    \caption{\textbf{Real-world task setups and qualitative comparison.} We evaluate our RL-finetuned Qwen3-VL (\ours) on toy arm repositioning, rope straightening, and cloth folding. 
    Leftmost columns show the initial and \textcolor[HTML]{9999FF}{target} states, with green and red arrows indicating predicted action vectors. 
    The time-lapse sequences contrast the successful executions of our \textcolor[HTML]{00A200}{RL-finetuned} model (top rows, green borders) against the typical failures of the pure \textcolor[HTML]{FF9999}{zero-shot} Qwen3-VL baseline (bottom rows, pink borders).}
    \label{fig:task_visualization}
    \vspace{-7mm}
\end{figure*}

\subsubsection{Automated Data Collection}
Training a video world model requires large-scale, diverse physical interaction data to accurately capture environment dynamics. While human teleoperation can provide high-quality demonstrations, acquiring large-scale datasets this way is difficult to scale. 
To improve collection efficiency for world model training, we develop an automated pipeline that leverages a zero-shot VLM to autonomously drive the data collection process.
By taking current and target RGB observations as input, the VLM maps visual goals to high-level action decisions, which are then executed through a specialized \emph{action primitive} to generate interactions without constant manual guidance.


\textbf{Action primitive.} We adopt discrete action commands as defined in ~\secref{subsection:vlm_planner}. Each action specifies a grasp location and a target placement location in the image plane. The selected pixel locations are transformed to 3D world coordinates and denoted as $p_{grasp}$ and $p_{target}$, respectively.

\begin{table}[b!]
    \vspace{-10pt}
    \centering
    \newcolumntype{C}{>{\centering\arraybackslash}X}
    \caption{\textbf{Ablation on Object-Only Video Prediction.} We compare the PSNR of synthetic rollouts generated by two action-conditioned video world models, trained for full-scene and object-only generation, respectively.}
    \label{tab:vdm}
    \begin{tabularx}{\linewidth}{lCCC}
        \toprule
         & Full-Scene Prediction & Object-Only Prediction \\
        \midrule
        PSNR $\uparrow$ & 24.70 & \textbf{26.25} \\
        \bottomrule
    \end{tabularx}
\vspace{-15pt}
\end{table}

\textbf{Pipeline Formulation.} 
At the beginning of the $k$-th action primitive, we observe the current image $I_k$ and aim to predict an action $a_k$ toward the goal image $I_g$ using a VLM.
However, since VLMs often struggle with precise spatial grounding~\cite{yang2023setofmarkpromptingunleashesextraordinary, OSWorld}, we first employ SAM2~\cite{ravi2024sam2segmentimages} to obtain accurate object masks. 
Specifically, the goal mask $M_g$ is pre-annotated offline. 
For the current scene, video SAM2 is initialized with a \emph{single} manual prompt(bounding box) provided once and then automatically propagates the mask, providing $M_k$ at the start of each action primitive.
To provide candidate locations for the VLM, we apply Farthest Point Sampling over $M_k$ and $M_g$ to extract point sets $\mathcal{P}_k$ and $\mathcal{P}_g$ from these masks. 
From these candidates, the VLM selects the arm $c$ and the specific point pair $(p_{grasp}, p_{target})$. 
Finally, these 2D coordinates are unprojected into 3D Cartesian waypoints for execution. The selected arm then follows a fixed motion sequence—moving from its home pose to the grasp point, closing the gripper, transporting the object to the target point, releasing, and returning to the home pose—with smooth trajectories generated via joint-space interpolation throughout. This cycle repeats iteratively to generate scalable exploratory training data.
Using this automated pipeline, we collect a total of 2056 trajectories 
for world model training.

\subsubsection{Evaluation Protocol}\label{subsubsec:evaluation_protocol}

To demonstrate the generalizability of our approach, we evaluate it on three representative real-world tasks requiring complex deformable object manipulation: rope straightening, cloth folding, and toy arm repositioning. \figref{fig:task_visualization} illustrates these task setups, contrasting the successful executions of our RL-finetuned model against typical zero-shot baseline failures.
In all tasks, the system is given a current observation image and a target image, and must predict actions to bring the object to the desired state. 
For each task, we conduct 10 rollout trials with randomized initial states and report the average score as the primary metric. 
In each trial, the system executes a single action primitive predicted from the current and target observation images, after which the resulting state is evaluated. 
Each trial is scored on a scale of $\{0, 0.5, 1\}$, where $1$ indicates complete success, $0.5$ indicates meaningful progress toward the target state, and $0$ indicates failure.

\subsection{Video World Models Serve as Reliable Verifiers}\label{subsec:vdm_as_verifier}

\subsubsection{Qualitative Comparison}
To assess whether video world models can faithfully simulate deformable dynamics and act as reliable verifiers, we qualitatively compare our method with two strong image-to-video baselines, CogVideoX-5B-I2V~\cite{yang2024cogvideox} and Wan2.2-I2V-A14B~\cite{wan2024wan2_2}. All models receive the same cropped object image as the initial frame. The baselines are conditioned via textual prompts describing grasp location and motion direction, while our model is conditioned on rendered robot arm trajectories.

As shown in \figref{fig:vdm_qualitative}, the baselines often fail to follow the specified actions and generate physically inconsistent deformations. In contrast, our action-conditioned world model produces motion that both adheres to the intended manipulation and exhibits physically plausible deformation behavior, making it reliable for VLM planner fine-tuning.

\subsubsection{Ablation on Object-Only Video Prediction}
We further investigate whether training the VDM to predict object-only videos improves modeling quality. As discussed in ~\secref{subsection:vlm_planner}, using cropped object videos encourages the model to focus on deformation dynamics rather than irrelevant visual details. To validate this design, we train two action-conditioned models under identical settings: one predicts full-scene videos including background and robot arms, while the other predicts cropped object-only videos rendered on a clean white background. Both models are conditioned on the same rendered robot arm trajectories. We compute the PSNR on the cropped object region in the final frame of the rollout. Results in \tabref{tab:vdm} show that the object-only model achieves higher PSNR, indicating more accurate reconstruction of deformable dynamics.



\subsection{World-Model RL Enhances VLM Planning}\label{subsec:wm_rl}

To answer \textbf{Q2}, we evaluate whether world-model-guided reinforcement learning enables the VLM planner to internalize physical dynamics. We first demonstrate that reinforcement learning significantly outperforms pure zero-shot planning. Specifically, our model, \ours, is developed by performing RL fine-tuning on the Qwen3-VL-8B base model. As illustrated in \figref{fig:zeroshot_vs_rl}, applying our RL pipeline yields a substantial improvement in task success rates compared to the zero-shot baseline. Across all evaluated tasks, we observe notable absolute gains ranging from 15\% to 40\%, intuitively highlighting how fine-tuning dramatically enhances the model's ability to successfully execute complex manipulation commands.

\begin{table}[t!]
    \centering
    \newcolumntype{C}{>{\centering\arraybackslash}X}
    \caption{\textbf{Inference Efficiency and Computational Cost Analysis.} We compare the task performance (Avg. Score), inference speed (Time), and computational overhead (TFLOPs) of \ours against a sampling-based explicit verification baseline. 
    }
    \label{tab:rl_inference_efficiency}
    \begin{tabularx}{\linewidth}{lCCC}
        \toprule
        \textbf{Method} & \textbf{Avg. Score $\uparrow$} & \textbf{Time (s) $\downarrow$} & \textbf{TFLOPs $\downarrow$} \\
        \midrule
        \textit{Explicit Verification} & & & \\
        \quad + Sample $N=4$   & 0.48 & 926.32 & $\text{7.82}\times \text{10}^{\text{4}}$ \\
        \quad + Sample $N=8$   & 0.50 & 2605.56 & $\text{1.56}\times \text{10}^{\text{5}}$ \\
        \midrule
        \textbf{\ours (Ours)} & \textbf{0.60} & \textbf{1.12} & \textbf{15.14} \\
        \bottomrule
    \end{tabularx}
    \vspace{-15pt}
\end{table}

To further validate this improvement, we quantitatively compare \ours against leading pure zero-shot baselines. As shown in \tabref{tab:rl_vs_zeroshot}, our method demonstrates a compelling advantage across all evaluated tasks involving complex deformable object manipulation. 
Notably, our method achieves an average score of 0.60, nearly doubling the overall performance of the strongest baseline (0.35). 
On specific tasks, \ours achieves 0.60 on Rope Straightening, doubling the performance of GPT-4o (0.30),and reaches 0.85 on Toy Arm Repositioning, outperforming the much larger Qwen3-VL-32B (0.70) by a considerable margin despite our model's smaller size.
This performance gap suggests that pure zero-shot VLMs, which rely primarily on visual features, struggle with the physical grounding required for dynamic manipulation. 
As shown in \figref{fig:task_visualization}, zero-shot actions often fail to align with the actual deformation behavior, a gap that our model bridges by internalizing physical dynamics.
By leveraging world model feedback from virtual rollouts to construct preference pairs, our RL training paradigm effectively bridges this gap, guiding the VLM to consistently favor physically viable strategies.

\subsection{Efficient Inference without World Model Sampling} 
To further contrast the efficiency of \ours, we evaluate a sampling-based baseline where the video world model acts as an explicit verifier.
Specifically, while \ours directly predicts physically grounded actions in a single forward pass, the sampling-based baseline operates differently:
Given the current observation and goal image, the zero-shot VLM proposes $N$ candidate actions. Each candidate is then rolled out by our video world model to generate a predicted future state, which is subsequently evaluated by GPT-4o to select the action most consistent with the target configuration. 
As detailed in \tabref{tab:rl_inference_efficiency}, while this explicit verification provides the necessary physical foresight to improve task performance, it incurs severe computational overhead.
The inference time increases with the number of sampled candidates $N$, requiring thousands of seconds per decision for only marginal performance gains.
In stark contrast, \ours achieves a superior average score while reducing the inference time by orders of magnitude, requiring only about $1$ seconds per decision.
\section{CONCLUSION}



We presented \ours, a novel framework that efficiently adapts vision-language planners for deformable manipulation by performing ORPO-based reinforcement learning entirely within the imagination of an action-conditioned video world model. Experiments on challenging deformable manipulation tasks show strong improvements over zero-shot planners without requiring large-scale real-world interaction.

\section{ACKNOWLEDGEMENT}
The USC Physical Superintelligence Lab acknowledges generous supports from Toyota Research Institute, Dolby, Google DeepMind, Capital One, Nvidia, and Qualcomm. Yue Wang is also supported by a Powell Research Award.

\bibliographystyle{ieeetr}  
\bibliography{ref}    

\end{document}